\documentclass[10pt,twocolumn,letterpaper]{article}

\usepackage[pagenumbers]{wacv} 

\definecolor{wacvblue}{rgb}{0.21,0.49,0.74}
\usepackage[pagebackref,breaklinks,colorlinks,allcolors=wacvblue]{hyperref}
\usepackage{tikz}
\usepackage{enumitem}
\usetikzlibrary{positioning, arrows.meta, shapes.multipart, calc}
\def\wacvPaperID{2624} 
\def\confName{WACV}
\def\confYear{2026}

\title{Morphing Through Time: Diffusion-Based Bridging of Temporal Gaps for Robust Alignment in Change Detection
}

\author{Seyedehnanita Madani \quad Vishal M.
Patel \\
Johns Hopkins University\\
{\tt\small smadani4@jhu.edu}  \quad {\tt\small vpatel36@jhu.edu }
}

\begin{document}
\maketitle

\begin{abstract}
Remote sensing change detection is often complicated by spatial misalignment between image pairs, especially when observations are separated by long temporal gaps such as seasonal or multi-year intervals. Conventional CNN- and transformer-based methods perform well on aligned data, but their reliance on perfect co-registration limits their applicability in practice. Existing approaches that integrate registration and change detection generally demand task-specific training and transfer poorly across domains.  We present a lightweight, modular pipeline that strengthens robustness without retraining the underlying change detection models. The framework combines rapid per-image LoRA adaptation with a compact flow refinement module trained under supervision. To mitigate large appearance differences, we generate intermediate morphing frames via a diffusion-based semantic interpolator. Consecutive frames are aligned using a registration backbone (e.g., RoMa), and the composed flows are further corrected through a residual refinement network. The refined flow is then applied to co-register the original image pairs, enabling more reliable downstream change detection. Extensive experiments on LEVIR-CD, DSIFN-CD, and WHU-CD demonstrate that the proposed pipeline significantly improves both registration accuracy and change detection performance, especially in scenarios with substantial spatial and temporal variations. 

\end{abstract}    

\section{Introduction}
\label{sec:intro}

Change detection (CD) in remote sensing underpins critical applications 
such as environmental monitoring, infrastructure mapping, urban 
development, and disaster assessment. The task involves identifying 
semantic differences between two images captured at different times. 
While modern deep learning models, ranging from CNNs~\cite{daudt2018fully} 
to transformers~\cite{bandara2022transformer}, perform well on aligned 
benchmarks, they assume pixel-level co-registration of the input pair. 
In practice, this assumption rarely holds.

Real-world satellite and aerial imagery often suffer from spatial 
misalignment caused by orbital drift, viewpoint variation, terrain 
parallax, or imperfect preprocessing. These misalignments are amplified 
over long temporal gaps, where seasonal change, illumination differences, 
and urban growth introduce substantial appearance shifts. As a result, 
state-of-the-art CD models degrade significantly under even modest 
misalignment, limiting their reliability in operational settings.

Dense registration methods aim to mitigate this by estimating pixel-wise 
flow or deformation fields. Recent advances such as 
RoMa~\cite{edstedt2024roma} and DVF-Net~\cite{xue2025dvf} leverage deep 
features to achieve high-quality matches. However, they still rely on 
photometric consistency and fail under severe domain shifts, where 
semantic changes dominate the visual signal. Directly applying these 
methods often produces unstable flows and weakens downstream CD.

Recent joint CD--registration frameworks~\cite{jing2025changerd} 
attempt to address this by sharing features across both tasks. While 
promising, they typically require large-scale annotated supervision, 
lack modularity, and often fail to generalize across sensors or unseen 
domains. 

Meanwhile, diffusion models have demonstrated strong semantic interpolation 
abilities. Prior CD work has explored diffusion for generating bi-temporal 
samples~\cite{tang2024changeanywhere}, latent interpolation~\cite{zheng2024changen2}, 
or learned priors~\cite{bandara2022ddpm, zhang2022smd}, but always under 
the assumption of aligned inputs. To our knowledge, no prior approach has 
combined diffusion-based semantic bridging with explicit registration and 
refinement for robust CD under misalignment.

We introduce a modular pipeline 
that combines diffusion-based morphing with dense registration and flow 
refinement. First, a 
DiffMorpher~\cite{zhang2024diffmorpher} module generates semantically 
coherent intermediate frames using lightweight, image-specific LoRA 
adapters. These intermediates transform a challenging long-range 
registration into a sequence of short-range problems. Second, 
RoMa~\cite{edstedt2024roma} estimates flows between consecutive morphs, 
which are composed into a global warp. Finally, a supervised 
\textbf{ResidualRefinerNet} corrects accumulated drift and recovers 
fine structures. The refined warp is applied to align bi-temporal pairs, 
which can then be processed by any frozen CD backbone.

Our pipeline is fully modular and general: morphing, registration, and 
refinement are decoupled, enabling integration with diverse CD models 
such as DDPM-CD~\cite{bandara2022ddpm}, ChangeFormer~\cite{bandara2022transformer}, 
and BIT-CD~\cite{chen2021remote}. We also benchmark against joint 
models like ChangeRD~\cite{jing2025changerd} (trained on our data) to 
ensure fairness.

\vspace{0.5em}
\noindent \textbf{Contributions.}  
\begin{itemize}[noitemsep,leftmargin=12pt]
    \item We propose a \textbf{modular CD pipeline} that integrates 
    \textbf{diffusion-based semantic morphing}, dense registration, 
    and residual refinement for robust alignment under severe temporal 
    domain shifts. 
    \item We design \textbf{ResidualRefinerNet}, a lightweight U-Net 
    that corrects accumulated drift in composed flows, significantly 
    improving alignment accuracy. 
    \item We conduct extensive experiments across four benchmarks 
    (LEVIR-CD, DSIFN-CD, WHU-CD, and S2Looking) and multiple CD 
    backbones (DDPM-CD, ChangeFormer, BIT-CD, ChangeRD), demonstrating 
    consistent gains in both registration quality and downstream CD.
\end{itemize}

\section{Related Work}
\label{sec:related}

Change detection in remote sensing has advanced rapidly with deep learning, yet most 
methods still assume that bi-temporal inputs are spatially aligned. In 
practice, misalignment from viewpoint differences, orbital drift, or 
terrain parallax is common, and is further compounded by appearance 
shifts over time (e.g., seasons, land use). Such inconsistencies 
degrade both registration and CD accuracy. We group related work into 
four areas.

\vspace{1mm}
\subsection*{Diffusion Models for CD}
Diffusion models have been applied to CD primarily for feature priors or 
data synthesis. DDPM-CD~\cite{bandara2022ddpm} leverages pretrained 
diffusion features to improve robustness under intra-class variation. 
SMDNet~\cite{zhang2022smd} injects diffusion priors into a Siamese 
encoder for better structural consistency. ChangeAnywhere~\cite{tang2024changeanywhere} 
and Changen2~\cite{zheng2024changen2} explore latent diffusion for 
temporal generation and augmentation. However, all assume aligned inputs. 
None use diffusion to explicitly address misalignment or improve 
correspondence.

\vspace{1mm}
\subsection*{Semantic Image Morphing}
Early morphing methods used mesh or field warping~\cite{beier2023feature, wolberg1998image}, 
while GAN-based approaches~\cite{isola2017image} interpolated in latent 
space but suffered from instability. Diffusion-based morphing has emerged 
as a stronger alternative. DiffMorpher~\cite{zhang2024diffmorpher} fits 
lightweight LoRA adapters to input images and interpolates latent noise 
and attention for semantically meaningful transitions. IMPUS~\cite{yang2023impus} 
uses diffusion control for identity-preserving translation. These methods 
focus on generative synthesis; none have been applied to registration or 
remote sensing CD. To our knowledge, diffusion-based semantic morphing 
has not previously been studied as an alignment strategy.

\vspace{1mm}
\subsection*{Dense Registration under Domain Shift}
Classical optical flow methods~\cite{horn1981determining, lucas1981iterative} 
rely on brightness constancy and fail under large shifts. Deep flow 
estimators such as PWC-Net~\cite{sun2018pwc}, RAFT~\cite{teed2020raft}, 
and D2-Net~\cite{dusmanu2019d2} improve robustness but still struggle 
with severe temporal variation. LoFTR~\cite{sun2021loftr} and MatchNet~\cite{han2015matchnet} 
remove explicit keypoint dependence, improving tolerance to distortion. 
RoMa~\cite{edstedt2024roma} combines DINOv2 features with a GP-based 
decoder for dense, keypoint-free matching and strong performance under 
moderate variation. Yet even RoMa degrades when semantic differences are 
large, motivating the need for additional guidance such as morphing-based 
intermediates.

\begin{figure*}[t]
  \centering
  \includegraphics[width=\textwidth]{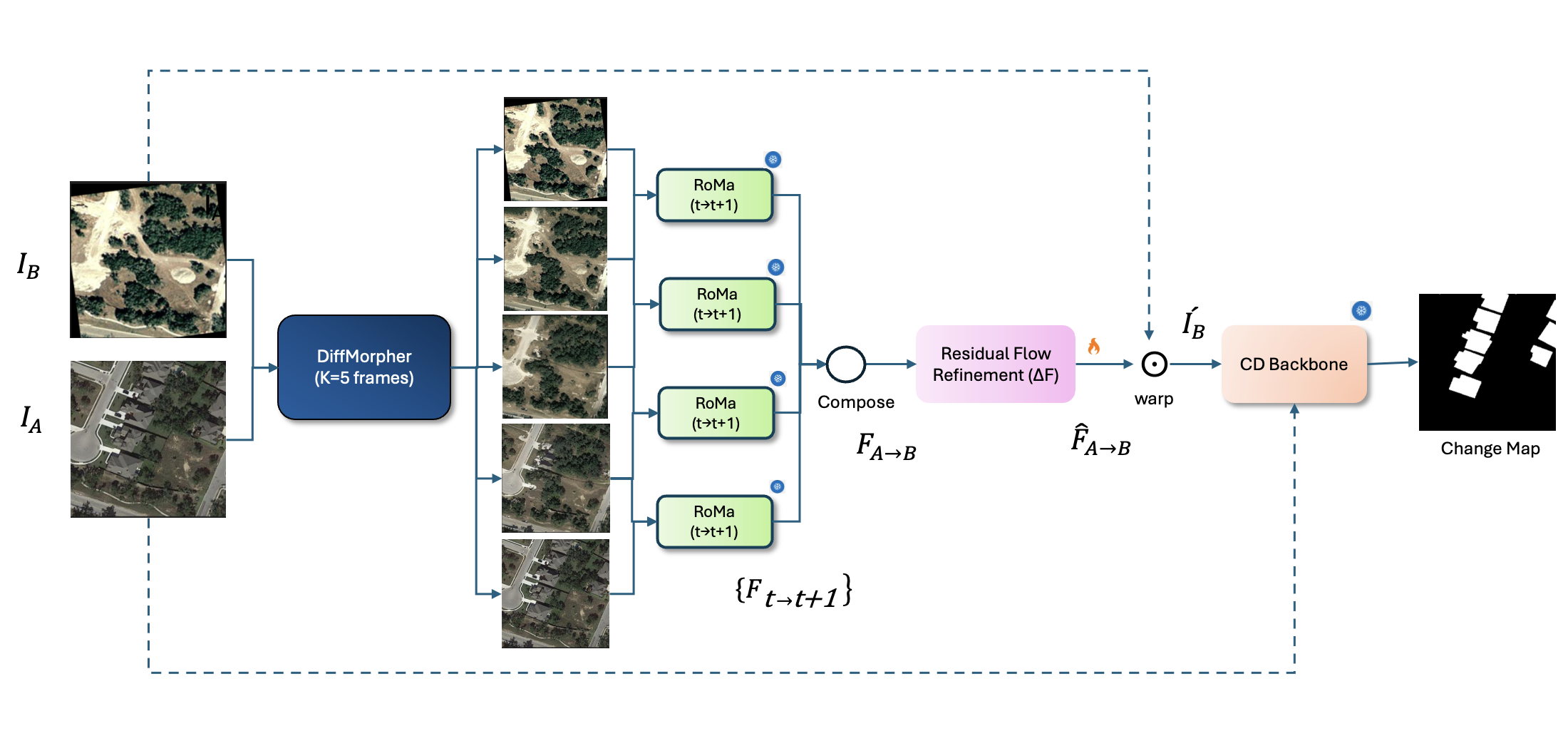}
 \vskip-25pt \caption{\textbf{Pipeline overview.} 
  Given bi-temporal images $I_A$ and $I_B$, \textbf{DiffMorpher} generates $K$ semantic intermediates. 
  \textbf{RoMa} estimates stepwise flows $\{F_{t \to t+1}\}$, which are \emph{composed} ($\circ$) into a global warp $F_{A \to B}$ 
  and refined by a \textbf{Residual Flow Refinement} module to $\hat{F}_{A \to B}$. 
  The refined flow \emph{warps} ($\odot$) $I_B$ to $I_B'$, and the pair $(I_A, I_B')$ is fed to a frozen CD backbone to produce the change map. 
  Dashed arrows denote raw inputs ($I_A, I_B$), while solid arrows denote intermediate signals. 
  (Illustrated with $K=5$.)}
  \label{fig:pipeline_overview}
\end{figure*}

\vspace{1mm}
\subsection*{Flow Refinement}
Flow refinement has been integrated into models like RAFT~\cite{teed2020raft} 
and GMA~\cite{xu2022gmflow}, where iterative correlation updates improve 
two-frame estimation. However, these refiners are tightly coupled to the 
backbone and not designed for composed, long-range flows. Our approach 
introduces a standalone \textbf{ResidualRefinerNet} that operates 
independently on composed flows, enabling post-hoc correction under large 
appearance changes without retraining the registration model.

\vspace{1mm}
\subsection*{Joint CD--Registration}
Several works attempt end-to-end joint solutions. ChangeRD~\cite{jing2025changerd} 
adds a geometric alignment module before CD prediction, while URCNet~\cite{zhou2023unified} 
and SimSaC~\cite{park2022dual} jointly optimize registration and change 
mask prediction. These methods improve robustness under mild 
misalignment, but require labeled supervision, retraining on each new 
domain, and offer limited modularity. Moreover, some (URCNet, SimSaC) 
lack public or complete code, limiting reproducibility. We therefore 
benchmark ChangeRD by retraining it on our perturbed data, and focus our 
comparisons on reproducible baselines with public code.

\vspace{1mm}
Prior work has explored diffusion for CD, morphing for synthesis, dense 
registration for correspondence, and joint models for CD+alignment. 
However, no method has combined diffusion-based semantic 
interpolation, stepwise flow composition, and 
standalone residual refinement in a modular pipeline. Unlike 
joint approaches, our method remains compatible with any frozen CD 
backbone (DDPM-CD, ChangeFormer, BIT-CD) and introduces supervision only 
in refinement, ensuring both practicality and generalization.

\begin{figure*}[t]
  \centering
  \textbf{Intermediate Frames} \\[0.5mm]
  \includegraphics[width=0.18\textwidth]{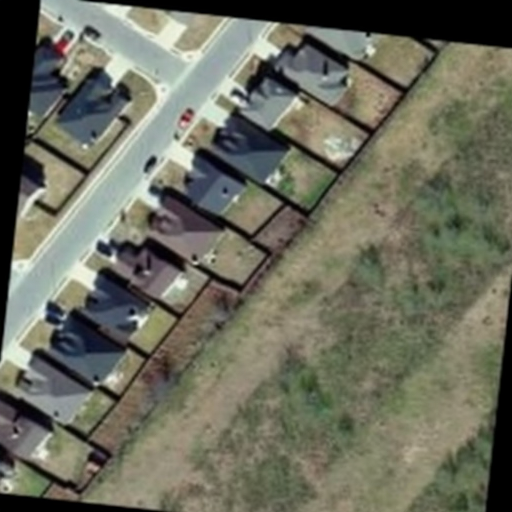}
  \includegraphics[width=0.18\textwidth]{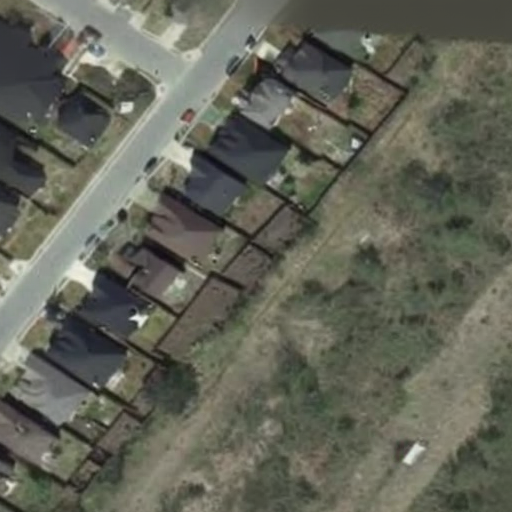}
  \includegraphics[width=0.18\textwidth]{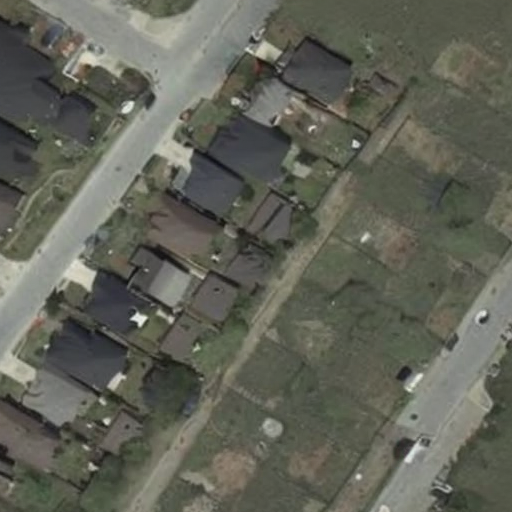}
  \includegraphics[width=0.18\textwidth]{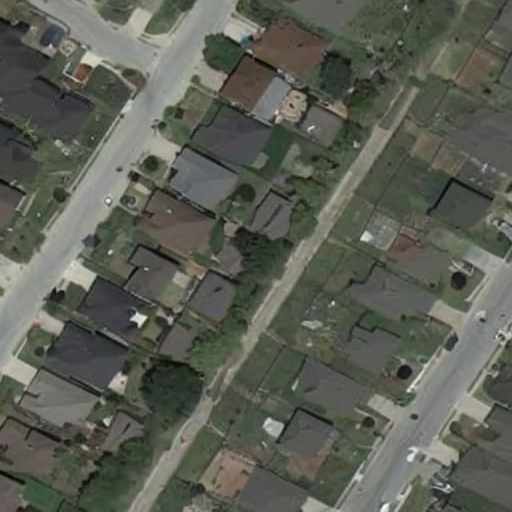}
  \includegraphics[width=0.18\textwidth]{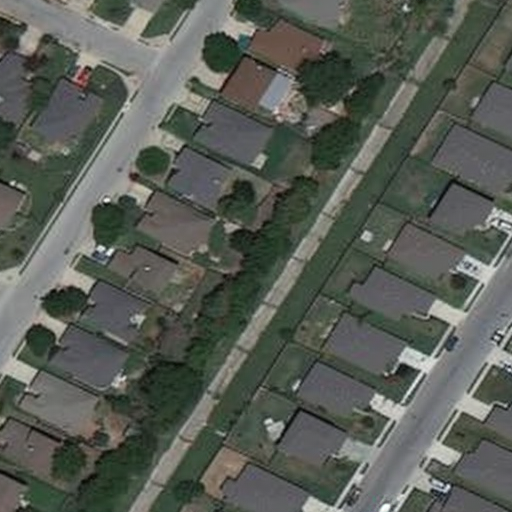}
  \\[1.5mm]
  \textbf{Warped Intermediates to $I_B$} \\[0.5mm]
  \includegraphics[width=0.18\textwidth]{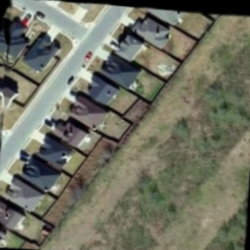}
  \includegraphics[width=0.18\textwidth]{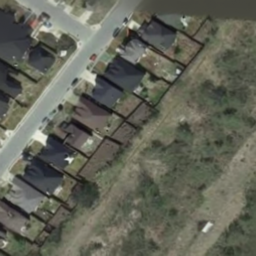}
  \includegraphics[width=0.18\textwidth]{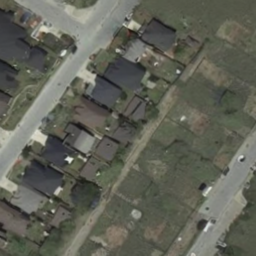}
  \includegraphics[width=0.18\textwidth]{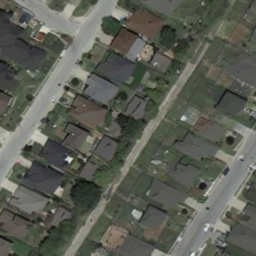}
  \includegraphics[width=0.18\textwidth]{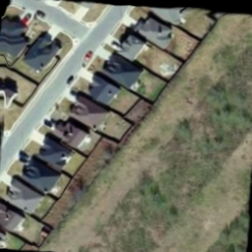}

  \caption{
    \textbf{Qualitative visualization of intermediate generation.}
    Top: Intermediate images generated via DiffMorpher. Bottom: their warped versions aligned to $I_B$. These sequential morphs enable more accurate motion decomposition and alignment across large scene shifts.
  }
  \label{fig:intermediate_warp}
\end{figure*}

\section{Methodology}
\label{sec:method}

\begin{itemize}
    \item \textbf{DiffMorpher:} We generate $K=5$ intermediate frames using image-specific LoRA tuning. Based on Zhang et al.~\cite{zhang2024diffmorpher}, LoRA fitting takes under 10 seconds per image, but this is a one-time offline cost amortized across morph pairs. Once tuned, morphing inference takes approximately \textbf{0.8 seconds} per image pair.
    
    \item \textbf{RoMa:} Dense flow estimation is performed between each morph pair. According to Edstedt et al.~\cite{edstedt2024roma}, RoMa inference runs at \textbf{0.2 seconds per pair}, totaling \textbf{1 second} for 5 morphs.
    
    \item \textbf{Flow Composition + Refinement:} Our ResidualRefinerNet adds \textbf{0.45 seconds} on average for full-resolution flow refinement and composition.
\end{itemize}


We propose a modular pipeline to address the challenges of CD under severe spatial and temporal misalignment (see Fig.~\ref{fig:pipeline_overview}). Large appearance shifts—caused by seasonal change, illumination variation, or urban growth—can severely degrade the performance of CD and registration models, especially when the image pairs are unaligned or visually dissimilar. Our pipeline tackles this by:  \\
(1) \textbf{Bridging semantic gaps} via intermediate frame synthesis with a diffusion-based morphing module;  \\
(2) \textbf{Stabilizing flow estimation} through multi-step dense registration and composition;  \\
(3) \textbf{Correcting spatial drift and refining detail} using a dedicated, trainable flow refinement network.  

This process produces well-aligned image pairs that can be used by off-the-shelf CD models without retraining, enabling robust and generalizable detection even under severe domain shifts.

\vspace{1mm}
\subsection{Problem Formulation}

Given a bi-temporal image pair $(I_A, I_B) \in \mathbb{R}^{H \times W \times 3}$ captured at times $t_A$ and $t_B$, the goal is to predict a binary change map $M_{AB} \in \{0,1\}^{H \times W}$ indicating semantic differences between the two images.  However, direct comparison of $(I_A, I_B)$ is unreliable due to significant appearance and viewpoint variations. To overcome this, we adopt a registration-based pipeline consisting of the following steps:

\begin{itemize}[noitemsep]
    \item \textbf{Semantic Morphing:} Generate a temporally coherent morphing sequence $\{I_t\}_{t=0}^K$ between $I_A$ and $I_B$ using a diffusion-based model. These intermediate frames enable finer-grained correspondence.
    
    \item \textbf{Local Flow Estimation:} Estimate dense optical flows $\{F_{t \rightarrow t+1}\}_{t=0}^{K-1}$ between each adjacent frame pair $(I_t, I_{t+1})$ in the morphing sequence.
    
    \item \textbf{Flow Composition:} Aggregate the local flows into a coarse global flow field $F_{A \rightarrow B}$, representing the estimated motion between $I_A$ and $I_B$.
    
    \item \textbf{Residual Refinement:} Refine $F_{A \rightarrow B}$ using a learned residual correction network to improve alignment accuracy.
    
    \item \textbf{Image Warping:} Apply the final refined flow to warp $I_B$ (or $I_A$), producing aligned inputs suitable for downstream CD models.
\end{itemize}

As illustrated in Fig. \ref{fig:intermediate_warp}, the intermediate frames and their warped counterparts facilitate accurate motion decomposition across large scene shifts. The aligned image pairs $(I_A, I_B')$ are passed to a frozen CD backbone. 
We evaluate multiple backbones, including 
DDPM-CD~\cite{bandara2022ddpm}, 
ChangeFormer~\cite{bandara2022transformerbasedsiamesenetworkchange}, 
and BIT-CD~\cite{Chen_2022}. 
All backbones are used with publicly available pretrained checkpoints 
and are never fine-tuned on perturbed data, ensuring that any 
performance improvements can be attributed solely to improved alignment.

\vspace{1mm}
\subsection{Semantic Morphing via DiffMorpher}

We adopt DiffMorpher~\cite{zhang2024diffmorpher}, a diffusion-based 
framework that generates high-fidelity semantic interpolations between 
input images. The model leverages Stable Diffusion with mechanisms to 
enforce coherence across interpolated frames. In our setup, we use the 
publicly available pretrained DiffMorpher backbone and apply only 
lightweight LoRA adapters, which are tuned per image pair. 
The core Stable Diffusion weights remain frozen, ensuring that 
adaptation is efficient and does not require re-training the full model.
This process yields semantically consistent intermediate frames that bridge IA and IB (see Fig.\ref{fig:intermediate_warp})
\begin{figure*}[t!]
  \centering
  \includegraphics[width=\linewidth, trim = 0 184pt 0 0, clip]{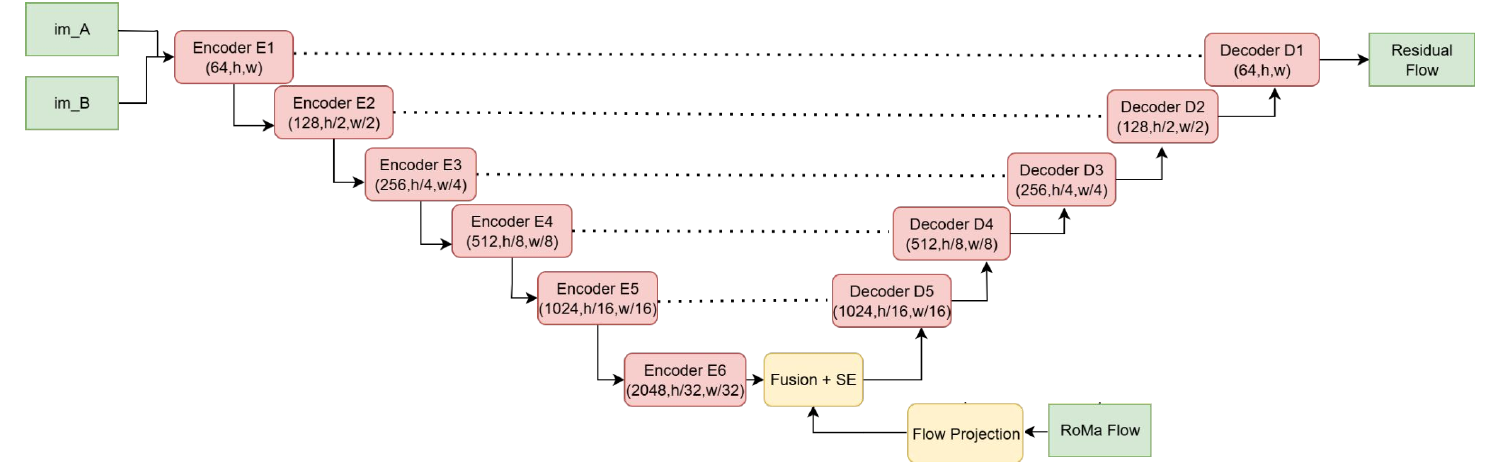}
  \caption{\textbf{ResidualRefinerNet architecture.} The input pair $(I_A, I_B)$ is encoded to a $32\times$ downsampled feature map. RoMa flow is projected and fused at the bottleneck. The decoder progressively upsamples and predicts residual flow $\Delta F$.}
  \label{fig:refiner_arch}
\end{figure*}

\paragraph{Overview.}
DiffMorpher fits lightweight LoRA~\cite{hu2022lora} modules to each input image, encoding high-level semantic identity. Intermediate transitions are synthesized by jointly interpolating the following components:

\begin{itemize}[noitemsep]
    \item \textbf{LoRA weights $\Delta\theta$:}  
    LoRA adapts pretrained Stable Diffusion weights by learning low-rank updates $\Delta\theta_A$ and $\Delta\theta_B$ for source and target images $I_A, I_B$. We linearly interpolate these updates:
    \begin{align*}
        \Delta\theta^{\alpha} = (1 - \alpha)\Delta\theta_A + \alpha\Delta\theta_B.
    \end{align*}

    \item \textbf{Latent noise vectors $z_T$:}  
    Each input is encoded into a diffusion latent with an associated terminal noise vector $z_T^A, z_T^B$. Following spherical interpolation, the blended latent noise at interpolation step $\alpha \in [0,1]$ is:
    \begin{align*}
        z_T^{\alpha} = \tfrac{\sin((1 - \alpha)\phi)}{\sin \phi} z_T^A + \tfrac{\sin(\alpha \phi)}{\sin \phi} z_T^B,
    \end{align*}
    where $\phi$ is the angle between $z_T^A$ and $z_T^B$.

    \item \textbf{Attention maps $(K,V)$:}  
    Stable Diffusion’s cross-attention uses key and value tensors $K,V$ extracted from each input. We interpolate them elementwise:
    \begin{align*}
        K^{\alpha} = (1 - \alpha)K^A + \alpha K^B, \quad
        V^{\alpha} = (1 - \alpha)V^A + \alpha V^B.
    \end{align*}

    \item \textbf{AdaIN statistics $(\mu,\sigma)$:}  
    Adaptive Instance Normalization modulates style using per-channel mean $\mu$ and variance $\sigma$. We interpolate these statistics:
    \begin{align*}
        \mu^{\alpha} = (1 - \alpha)\mu_A + \alpha\mu_B, \quad
        \sigma^{\alpha} = (1 - \alpha)\sigma_A + \alpha\sigma_B.
    \end{align*}
\end{itemize}
The interpolated components are injected into the pretrained UNet of Stable Diffusion to denoise the latent and synthesize semantically consistent intermediate images.

\paragraph{Morph Sequence.}
Given $I_A$ and $I_B$, the model produces a sequence of $K$ intermediate images:
\[
I_A = I_0, I_1, \dots, I_K = I_B.
\]
We use $K = 5$ in our experiments to balance granularity and efficiency. This morphing provides a continuous trajectory from source to target, which we exploit for dense correspondence estimation.

\vspace{1mm}
\subsection{Multi-step Dense Registration via RoMa}

We adopt RoMa~\cite{edstedt2024roma} to compute dense correspondences between consecutive morph frames $(I_t, I_{t+1})$. RoMa is a coarse-to-fine matcher that outputs a dense flow field $\hat{F}_{t \rightarrow t+1}$ and a per-pixel certainty map. Its pipeline can be summarized as follows.

\paragraph{1. Feature Extraction.}

RoMa extracts coarse features from a frozen DINOv2 backbone (stride 14) and fine features from a VGG19 encoder:
\[
\{\phi_A^{\text{coarse}}, \phi_A^{\text{fine}}\} = F_\theta(I_A), \quad 
\{\phi_B^{\text{coarse}}, \phi_B^{\text{fine}}\} = F_\theta(I_B).
\]

\paragraph{2. Cosine Similarity + GP Encoding.}

Normalized coarse features form a cosine similarity matrix:
\[
s_{ij} = \frac{\langle \phi_A^{\text{coarse}}(i), \phi_B^{\text{coarse}}(j) \rangle}
{\|\phi_A^{\text{coarse}}(i)\| \cdot \|\phi_B^{\text{coarse}}(j)\|}.
\]
A Gaussian Process module smooths this map to yield a probabilistic match embedding $z_{\text{GP}}$.

\paragraph{3. Transformer Decoder.}

$z_{\text{GP}}$ and DINOv2 features are passed to a Transformer decoder that predicts match distributions over anchor locations. The coarse warp is recovered via soft-argmax over the most likely anchors.

\paragraph{4. Multi-scale Refinement.}

The coarse warp is refined across scales $\{8,4,2,1\}$ using convolutional refiners $R_{\theta,s}$:
\[
\hat{F}_s = \text{Upsample}(\hat{F}_{s+1}) + \Delta F, \quad 
p_s = \text{Upsample}(p_{s+1}) + \Delta p.
\]
This produces a high-resolution flow $\hat{F}_{A \rightarrow B}$ and certainty map $p(x_A)$.

\paragraph{Flow Composition.}

RoMa is applied to each morph pair to obtain short-range flows $\{F_{t \rightarrow t+1}\}_{t=0}^{K-1}$, which are composed as:
\begin{align}
F_{A \rightarrow B} &= F_{0 \rightarrow 1} \oplus F_{1 \rightarrow 2} \oplus \cdots \oplus F_{K-1 \rightarrow K}, \\
(F \oplus G)(x) &= F(x) + G(x + F(x)).
\end{align}
The composed flow captures dense pixel-level correspondences from $I_A$ to $I_B$ even under large temporal or semantic shifts (Fig.~\ref{fig:flow_refine}).

\subsection{Flow Refinement}

To address flow refinement, we propose \textbf{ResidualRefinerNet}, a dedicated trainable module designed to correct accumulated drift and recover fine structural details in the composed flow.  This is the only trainable component in our pipeline and plays a central role in our performance gains. Figure \ref{fig:refiner_arch} provides an overview of the ResidualRefinerNet architecture, which refines the composed flow by correcting accumulated errors.

\vspace{1mm}
Unlike traditional optical flow refinement that relies only on photometric consistency, our goal is to enhance geometric alignment in a way that benefits semantic tasks like CD. The refiner uses both image context and initial flow estimates to resolve subtle misalignments, particularly at object boundaries and under appearance change.

\vspace{2mm}

To correct cumulative misalignments in the composed flow $F_{A \rightarrow B}$, we introduce \textbf{ResidualRefinerNet}, a U-Net-style residual flow correction network. It refines coarse global flow using joint reasoning over the source image $I_A$, target image $I_B$, and the input flow field.

\begin{figure}[t]
  \centering
  \textbf{Final Warps} \\[0.5mm]
  \includegraphics[width=0.31\linewidth]{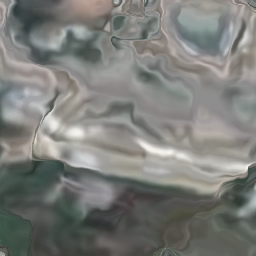}
  \includegraphics[width=0.31\linewidth]{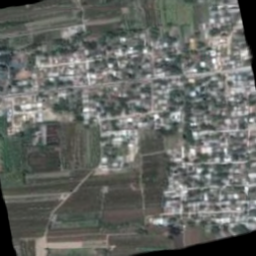}
  \includegraphics[width=0.31\linewidth]{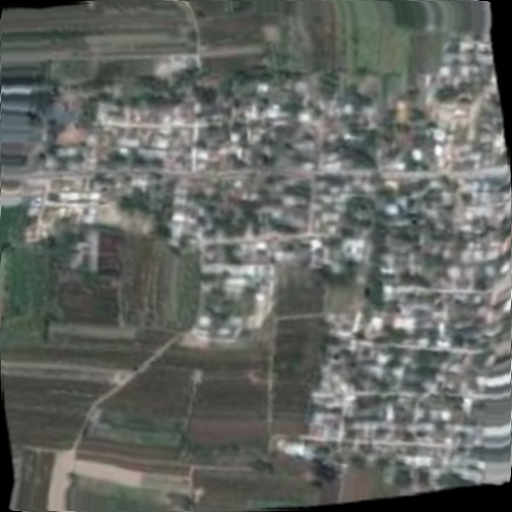}
  \\[-0.4em]
  {\small Direct \hspace{3em} Composed (Morph only) \hspace{3em} Refined}
  \\[1.5mm]
  \textbf{Flow Maps} \\[0.5mm]
  \includegraphics[width=0.31\linewidth]{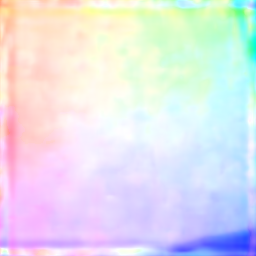}
  \includegraphics[width=0.31\linewidth]{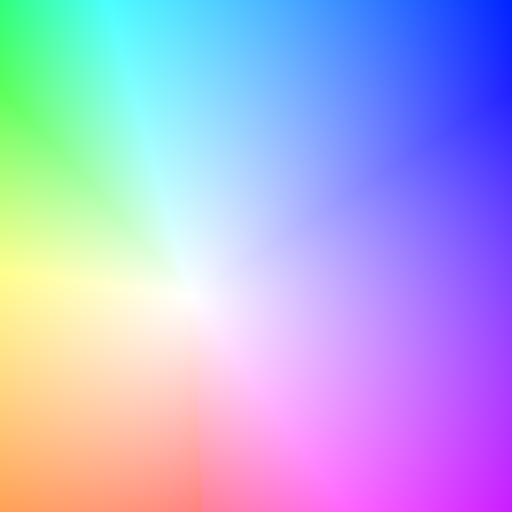}
  \includegraphics[width=0.31\linewidth]{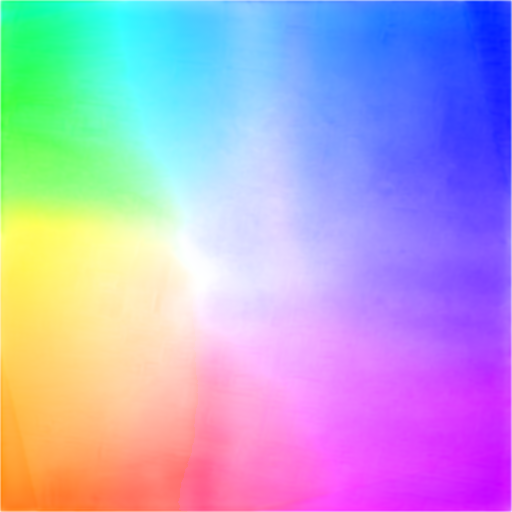}
  \\[-0.4em]
  {\small Input \hspace{3em} Ground Truth \hspace{3em} Refined}

  \caption{
    \textbf{Final alignment and flow visualization.}
    Top: composed vs. direct RoMa warp and our refined result. Bottom: corresponding flow maps (input, GT, refined) reveal improvements in structure and smoothness.
  }
  \label{fig:flow_refine}
\end{figure}

The input tensor is defined as:
\[
X = \text{concat}(I_A, I_B, F_{A \rightarrow B}) \in \mathbb{R}^{(6 + 2) \times H \times W},
\]
where $I_A, I_B \in \mathbb{R}^{3 \times H \times W}$ are RGB images and $F_{A \rightarrow B} \in \mathbb{R}^{2 \times H \times W}$ is the coarse flow. The concatenated tensor $X$ is processed as follows:

\begin{itemize}[noitemsep]
    \item \textbf{Encoder.} A series of convolutional blocks $\{E_i\}_{i=1}^6$ extract multi-scale visual features:
    \[
    f_i = E_i(f_{i-1}), \quad f_0 = X,
    \]
    where $f_i \in \mathbb{R}^{C_i \times H_i \times W_i}$ denotes features at resolution level $i$.

   \item \textbf{Flow Fusion at Bottleneck.} The original flow $F_{A \rightarrow B}$ is bilinearly downsampled to match the bottleneck resolution $(H_6, W_6)$ and projected to match the channel dimension via a $1 \times 1$ convolution:
\[
\tilde{F} = \phi(F_{A \rightarrow B}) \in \mathbb{R}^{2048 \times H_6 \times W_6}.
\]
The projected flow is then added element-wise to the bottleneck feature $f_6$, and the result is passed through a squeeze-and-excitation (SE) module $\mathcal{S}$:
\[
z = \mathcal{S}(f_6 + \tilde{F}).
\]

    \item \textbf{Decoder.} Features are progressively upsampled and combined with encoder features via skip connections:
    \[
    \hat{f}_{i} = D_i\left( \text{concat}( \text{upsample}(\hat{f}_{i+1}), f_i ) \right), \quad i=5,\dots,1,
    \]
    where $D_i$ denotes the $i$-th decoder block, and $\hat{f}_1$ is the final decoded feature.

    \item \textbf{Residual Flow Head.} A final $3 \times 3$ convolution head maps $\hat{f}_1$ to the residual flow correction:
    \[
    \Delta F = \psi(\hat{f}_1) \in \mathbb{R}^{2 \times H \times W}.
    \]
    The final refined flow is obtained via:
    \[
    \hat{F}_{A \rightarrow B} = F_{A \rightarrow B} + \Delta F.
    \]
\end{itemize}
\paragraph{Loss Function.}
The network is trained using a pixel-wise Smooth L1 loss between the refined flow $\hat{F}_{A \rightarrow B}$ and the ground truth flow $F^*_{A \rightarrow B}$:
\[
\mathcal{L}_{\text{refine}} = \frac{1}{|\Omega|} \sum_{x \in \Omega} \text{SmoothL1} \left( \hat{F}(x) - F^*(x) \right),
\]
wheere,
\[
\text{SmoothL1}(e) =
\begin{cases}
0.5 e^2, & \text{if } |e| < 1 \\
|e| - 0.5, & \text{otherwise}.
\end{cases}
\]
This formulation ensures robustness to large outliers and stabilizes learning in challenging regions such as occlusions, textureless surfaces, or morph-induced distortions.
Figure \ref{fig:flow_refine} visualizes the final flow fields and their corresponding warps, revealing structural improvements and smoothness in the refined results.

\section{Experiments}
\label{sec:experiments}

\begin{table}[h!]
\centering
\normalsize   
\begin{tabular}{lccc}
\toprule
\textbf{Dataset} & \textbf{Metric} & \textbf{Composed} & \textbf{Direct} \\
\midrule
LEVIR  & PSNR$\uparrow$ & 30.53 & 20.93 \\
       & SSIM$\uparrow$ & 0.9318 & 0.5117 \\
\midrule
WHU    & PSNR$\uparrow$ & 27.98 & 19.20 \\
       & SSIM$\uparrow$ & 0.9069 & 0.4605 \\
\midrule
DSIFN  & PSNR$\uparrow$ & 33.39 & 23.15 \\
       & SSIM$\uparrow$ & 0.9516 & 0.5915 \\
\bottomrule
\end{tabular}
\caption{Image quality results comparing Direct and Composed flows.}
\label{tab:img_quality}
\end{table}

\vspace{-1mm}
\begin{figure*}[t]
  \centering
  \includegraphics[width=0.19\linewidth]{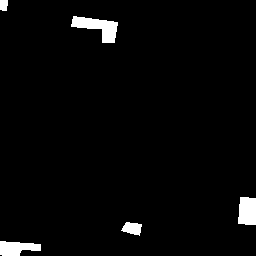}
  \includegraphics[width=0.19\linewidth]{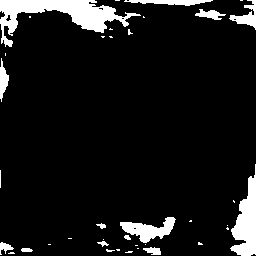}
  \includegraphics[width=0.19\linewidth]{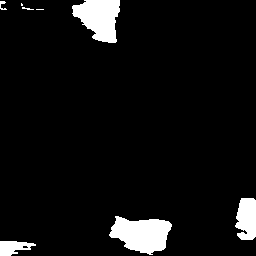}
  \includegraphics[width=0.19\linewidth]{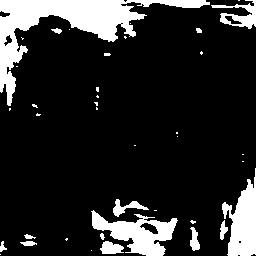}
  \includegraphics[width=0.19\linewidth]{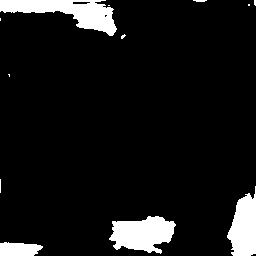}

  \caption{
    \textbf{CD outputs.}
    From left to right: ground truth mask, prediction from unaligned pair, and predictions after alignment using direct RoMa, composed RoMa, and refined RoMa.
  }
  \label{fig:change_masks_pair2}
\end{figure*}

\begin{figure*}[t]
  \centering
  \includegraphics[width=0.19\linewidth]{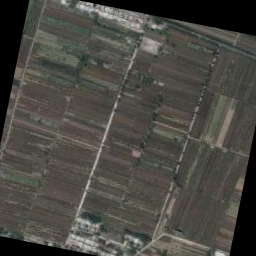}
  \includegraphics[width=0.19\linewidth]{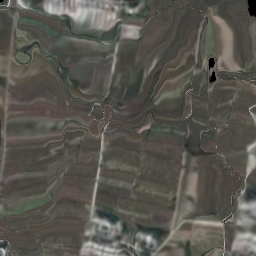}
  \includegraphics[width=0.19\linewidth]{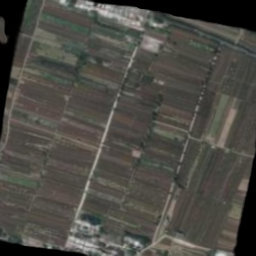}
  \includegraphics[width=0.19\linewidth]{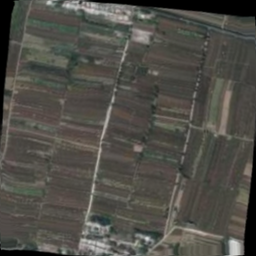}
  \includegraphics[width=0.19\linewidth]{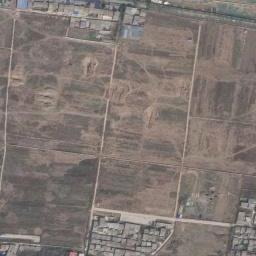}

  \caption{
    \textbf{Alignment stages }
    The first four images show unaligned $I_A$ and three alignment outputs (Direct, Composed (morph-only), Refined). The last image is $I_B$ (target image).
  }
  \label{fig:intermediate_pair2}
\end{figure*}

We evaluate our pipeline on three widely used change detection (CD) datasets: \textbf{LEVIR-CD} (637 urban scenes with multi-year building changes), \textbf{WHU-CD} (1,368 pairs capturing urban expansion under diverse viewpoints and seasonal shifts), and \textbf{DSIFN-CD} (1,084 pairs from multiple cities with fine-grained changes). All datasets are originally well aligned; synthetic perturbations are introduced only to enable controlled evaluation and training of our refinement module. Importantly, CD backbones are never retrained on perturbed data.

Because standard CD datasets are already co-registered, we simulate misalignment using random affine transformations (translation, rotation, scaling) applied to one image in each pair. This generates controlled distortions while providing exact pixel-level flow supervision at $256 \times 256$. Perturbations are applied only to the test split, which is then subdivided into 80\%/10\%/10\% train/val/test for refinement training. CD models are always frozen and evaluated in zero-shot mode.

\vspace{1mm}
\noindent\textbf{Evaluation Metrics.}  
We adopt a dual protocol covering semantic segmentation and alignment accuracy.  
\textit{CD metrics} include F1-scores (change / no-change), their average (mF1), mean IoU (mIoU), and overall accuracy (OA).  
\textit{Alignment metrics} include End-Point Error (EPE, lower is better) and Enhanced Correlation Coefficient (ECC), defined as $1-\cos(\theta)$ between warped and target images (lower indicates better global alignment). This combination measures both local flow accuracy and global consistency. As shown in Table \ref{tab:img_quality}, morph-based composition substantially improves image fidelity, boosting PSNR by 9–10 dB and SSIM by +0.40 on average across datasets compared to direct RoMa.
\textit{Image quality metrics} (for warped reconstructions) include PSNR and SSIM, which capture perceptual fidelity of aligned pairs.

\vspace{1mm}

\vspace{1mm}
\noindent\textbf{Alignment Variants.} We benchmark six settings:
\begin{enumerate}[labelindent=20pt]
    \item \textbf{Original} — perfectly aligned pairs;
    \item \textbf{Unaligned} — perturbed pairs without correction;
    \item \textbf{Dir (RoMa)} — single-step dense flow from RoMa;
    \item \textbf{Dir+Refined} — RoMa direct flow corrected with our ResidualRefinerNet;
    \item \textbf{Composed} — sequential RoMa flows across DiffMorpher-generated morphs;
    \item \textbf{Composed+Refined (Ours)} — final variant with residual refinement.
\end{enumerate}

\vspace{1mm} 
Table~\ref{tab:reg_backbones} compares dense registration baselines. RAFT achieves very low EPE but much higher ECC (e.g., 23.0/1.58 on LEVIR), suggesting poor global alignment under temporal shifts. RoMa produces lower ECC while preserving competitive EPE, and our refinement stage reduces both simultaneously (e.g., 3.41/0.035 on LEVIR). We therefore adopt RoMa as the primary registration backbone in all experiments.


We first quantify the effect of alignment on image similarity (Table~\ref{tab:img_quality}). 
The Composed (morph-only) variant achieves perceptually closer reconstructions than Direct RoMa, improving PSNR by +9--10 dB and SSIM by +0.40 on average across LEVIR, WHU, and DSIFN. 
For example, on LEVIR, Composed reaches 30.53 dB / 0.93 SSIM compared to 20.93 dB / 0.51 SSIM for Direct. 
These results confirm that semantic morphing produces more faithful interpolations, providing a stronger initialization for refinement.
These quantitative gains are reflected in qualitative masks (Fig. \ref{fig:change_masks_pair2}), where refined flows reduce false positives and sharpen building boundaries.
We then evaluate downstream CD performance using three state-of-the-art backbones in frozen inference mode: 
DDPM-CD~\cite{bandara2022ddpm}, ChangeFormer~\cite{bandara2022transformerbasedsiamesenetworkchange}, and BIT-CD~\cite{Chen_2022}. 
All are applied with publicly released checkpoints (e.g., 50/100/400 steps for DDPM-CD) and are never fine-tuned on perturbed data, ensuring that any differences arise solely from alignment. 
Figure~\ref{fig:intermediate_pair2} illustrates qualitative gains, while Tables~\ref{tab:levir_flow_compact}, \ref{tab:whu_flow_compact} and \ref{tab:dsifn_flow_compact} show consistent quantitative improvements. 
As summarized in Table \ref{tab:cd_backbones_under_alignment}, across all three datasets our refined alignment consistently boosts performance regardless of backbone, demonstrating that the pipeline is model-agnostic

\begin{table}[h!]
\centering
\scriptsize
\resizebox{\columnwidth}{!}{
\begin{tabular}{lcccc}
\toprule
\textbf{Experiment} & \textbf{mF1 $\uparrow$} & \textbf{mIoU $\uparrow$} & \textbf{OA $\uparrow$} & \textbf{F1$_1$  $\uparrow$} \\
\midrule
Composed \& Refined Flow & \textbf{.93} & \textbf{.88} & \textbf{.98} & \textbf{.87} \\
Direct Refined Flow      & .92 & .86 & .97 & .85 \\
Unaligned                & .91 & .85 & .98 & .83 \\
\textit{ChangeRD}$^\dagger$ & .92 & .88 & .98 & .87 \\
Composed                 & .91 & .85 & .98 & .84 \\
Dir Warped               & .90 & .83 & .98 & .81 \\
Original                 & \underline{.95} & \underline{.91} & \underline{.99} & \underline{.91} \\
\bottomrule
\end{tabular}
}
\caption{LEVIR-CD results using RoMa-based alignment. $^\dagger$ChangeRD baseline.}
\label{tab:levir_flow_compact}
\end{table}

\begin{table}[h!]
\centering
\scriptsize
\resizebox{\columnwidth}{!}{
\begin{tabular}{lcccc}
\toprule
\textbf{Experiment} & \textbf{mF1  $\uparrow$} & \textbf{mIoU  $\uparrow$} & \textbf{OA  $\uparrow$} & \textbf{F1$_1$  $\uparrow$} \\
\midrule
Composed \& Refined Flow & \textbf{.82} & \textbf{.72} & \textbf{.97} & \textbf{.65} \\
Direct Refined Flow      & .79 & .69 & .96 & .60 \\
Unaligned                & .75 & .65 & .94 & .52 \\
\textit{ChangeRD}$^\dagger$ & .70 & .61 & .93 & .45 \\
Composed                 & .79 & .69 & .96 & .60 \\
Dir Warped               & .62 & .54 & .93 & .28 \\
Original                 & \underline{.81} & \underline{.72} & \underline{.97} & \underline{.64} \\
\bottomrule
\end{tabular}
}
\caption{WHU-CD results using RoMa-based alignment. $^\dagger$ChangeRD baseline.}
\label{tab:whu_flow_compact}
\end{table}

\begin{table}[h!]
\centering
\scriptsize
\resizebox{\columnwidth}{!}{
\begin{tabular}{lcccc}
\toprule
\textbf{Experiment} & \textbf{mF1  $\uparrow$} & \textbf{mIoU  $\uparrow$} & \textbf{OA  $\uparrow$} & \textbf{F1$_1$  $\uparrow$} \\
\midrule
Composed \& Refined Flow & \textbf{.90} & \textbf{.83} & \textbf{.94} & \textbf{.84} \\
Direct Refined Flow      & .90 & .81 & .94 & .83 \\
Unaligned                & .89 & .81 & .93 & .82 \\
\textit{ChangeRD}$^\dagger$ & .45 & .42 & .83 & .71 \\
Composed                 & .89 & .80 & .93 & .82 \\
Dir Warped               & .76 & .63 & .83 & .64 \\
Original                 & \underline{.96} & \underline{.92} & \underline{.97} & \underline{.93} \\
\bottomrule
\end{tabular}
}
\caption{DSIFN-CD results using RoMa-based alignment. $^\dagger$ChangeRD baseline.}
\label{tab:dsifn_flow_compact}
\end{table}

\begin{table}[t]
\centering
\setlength{\tabcolsep}{3pt} 
\scriptsize
\resizebox{\columnwidth}{!}{
\begin{tabular}{lccc}
\toprule
\textbf{Registration} & \textbf{LEVIR} & \textbf{WHU} & \textbf{DSIFN} \\
 & (ECC $\downarrow$/EPE$\downarrow$) & (ECC$\downarrow$/EPE$\downarrow$) & (ECC$\downarrow$/EPE$\downarrow$) \\
\midrule
SP{+}SG & .23 / 308.27 & .38 / 243.96 & .08 / 397.21 \\
LoFTR   & .24 / 229.34 & .34 / 217.52 & .10 / 275.00 \\
RAFT    & 1.58 / 23.03  & 1.63 / 24.65  & 1.27 / 17.25 \\
MASt3R  & .48 / 57.60 & .43 / 53.84 & .67 / 82.66 \\   
RoMa (Dir)       & .89 / 29.41 & .87 / 25.78 & .99 / 21.76 \\
RoMa (Composed)  & .99 / 2.94  & 1.02 / 2.76  & .92 / 2.93  \\
\textbf{RoMa + Refined (Ours)} & \textbf{.04 / 3.41} & \textbf{.06 / 4.08} & \textbf{.05 / 4.03} \\
\bottomrule
\end{tabular}
}
\caption{Registration-only comparison across backbones.}

\label{tab:reg_backbones}
\end{table}

\begin{table}[t]
\centering
\scriptsize
\setlength{\tabcolsep}{3pt}
\resizebox{\columnwidth}{!}{%
\begin{tabular}{lcccc}
\toprule
\textbf{CD Backbone} & \textbf{Unaligned} & \textbf{Dir} & \textbf{Composed} & \textbf{Refined (Ours)} \\
\midrule
\multicolumn{5}{c}{\textit{LEVIR-CD}} \\
ChangeFormer & .91/.85/.98 & .92/.86/.98 & .91/.85/.99 & \textbf{.93/.88/.99} \\
BIT\text{-}CD   & .91/.85/.98 & .92/.86/.98 & .91/.85/.98 & \textbf{.93/.88/.99} \\
DDPM\text{-}CD & .91/.85/.98 & .90/.84/.98 & .92/.85/.99 & \textbf{.93/.88/.99} \\
\midrule
\multicolumn{5}{c}{\textit{DSIFN-CD}} \\
ChangeFormer & .82/.71/.88 & .76/.64/.85 & .85/.74/.90 & \textbf{.87/.78/.92} \\
BIT\text{-}CD   & .45/.40/.81 & .45/.41/.81 & .45/.41/.81 & \textbf{.45/.41/.81} \\
DDPM\text{-}CD & .89/.81/.93 & .76/.63/.83 & .89/.80/.93 & \textbf{.90/.83/.94} \\
\midrule
\multicolumn{5}{c}{\textit{WHU-CD}} \\
BIT\text{-}CD   & .50/.48/.96 & .49/.48/.96 & .50/.48/.96 & \textbf{.50/.48/.95} \\
DDPM\text{-}CD & .75/.65/.95 & .62/.54/.92 & .79/.69/.96 & \textbf{.79/.69/.96} \\
\bottomrule
\end{tabular}
}
\caption{CD backbones under identical alignment settings (entries are mF1$\uparrow$/mIoU$\uparrow$/OA$\uparrow$; CD is zero-shot).}
\label{tab:cd_backbones_under_alignment}
\end{table}


Overall, the modular pipeline achieves inference in roughly \textbf{2.7 seconds per image pair}, and its design allows for parallelized morphing and registration stages. Unlike retraining-heavy approaches such as URCNet, our system offers plug-and-play efficiency and generalizability without model fine-tuning.

\section{Conclusion}
\label{sec:conclusion}

We presented a modular pipeline for robust change detection (CD) under severe spatial and temporal misalignment. By combining diffusion-based semantic morphing, dense registration with RoMa, and residual flow refinement, our framework provides a plug-and-play alignment solution without retraining CD models.

Experiments on LEVIR-CD, WHU-CD, and DSIFN-CD show that misalignment sharply degrades CD accuracy, while our method consistently mitigates this effect. As shown in Fig. \ref{fig:change_masks_pair2}, alignment substantially improves the quality of CD outputs, underscoring the practical impact of our pipeline. The refined flow variant reduces registration error by up to 90\% (EPE) and raises mean F1 scores by as much as +5 points, confirming both accuracy and generality.

\textbf{Limitations and Future Work.} Current training uses synthetic affine perturbations, which may not capture complex real-world distortions. Future directions include unsupervised or weakly supervised refinement (e.g., photometric or cycle consistency) and evaluation on broader datasets.

\textbf{Takeaway.} Alignment remains a key bottleneck in CD. Our diffusion-bridged, refinement-based pipeline offers a practical, model-agnostic solution with immediate impact on real-world remote sensing workflows.

{
    \small
    \bibliographystyle{ieeenat_fullname}
    \bibliography{main}
}

\end{document}